# Comparing Synthetic Tabular Data Generation Between a Probabilistic Model and a Deep Learning Model for Education Use Cases

Herkulaas MvE Combrink [1] [0000-0001-7741-3418], Vukosi Marivate [1] [0000-0002-6731-6267] and Benjamin Rosman [2] [0000-0002-0284-4114]

[1] Department of Computer Science, University of Pretoria, South Africa
[2] School of Computer Science and Applied Mathematics, University of the Witwatersrand, Johannesburg, South Africa
u29191051@tuks.co.za

**Abstract.** The ability to generate synthetic data has a variety of use cases across different domains. In education research, there is a growing need to have access to synthetic data to test certain concepts and ideas. In recent years, several deep learning architectures were used to aid in the generation of synthetic data – but with varying results. In the education context, the sophistication of implementing different models requiring large datasets is becoming very important. This study aims to compare the application of synthetic tabular data generation between a probabilistic model specifically a Bayesian Network, and a deep learning model, specifically a Generative Adversarial Network using a classification task. The results of this study indicate that synthetic tabular data generation is better suited for the education context using probabilistic models (overall accuracy of 75%) than deep learning architecture (overall accuracy of 38%) because of probabilistic interdependence. Lastly, we recommend that other data types, should be explored and evaluated for their application in generating synthetic data for education use cases.

## 1 Introduction

The ability to generate synthetic data has a variety of use cases across different domains [1, 2]. Traditionally, synthetic data generation was computationally implemented using different types of probabilistic models [3, 4, 5]. However, in recent years, several deep learning architectures were used to aid in the generation of synthetic data – but with varying results [6, 7]. The need for synthetic data generation is becoming more important. This is, firstly, because large training datasets are hard to come by for every specific use case. Secondly, the implementation of different models requires large datasets to examine its efficacy [8, 9]. This study aims to compare the application of syn-thetic tabular data generation between a Bayesian Network (BN), and a Generative Adversarial Network (GAN) using descriptive statistics, and an evaluation of machine learning accuracy scores [10 - 13]. The ability to create usable synthetic data from a small sample of information is of growing importance in a variety of different domains such as medicine, education, engineering, language, and business - to name a few. Furthermore, once the correct models have been created to generate synthetic data for a specific use case, the costs of running experiments and simulations are reduced because the data can be generated in place of such experimentation [14, 15]. Within the context of higher education, finding relevant data remains a challenge [16]. These challenges specifically relate to the way in which the data is shared, and less with the availability of the data itself, as such, generating synthetic data for higher education is of importance for higher education use cases. A lot of this information is driven by the need to improve the accuracy of digital twin technology. Digital twinning refers to a virtual equivalent of a realworld application or system that functions on the same logic and rules of nature as the physical counterpart [17, 18]. An application of digital twin for education could mean the generation of infinite student data within a higher education simulation. It is estimated that digital twin technology will drastically reduce the cost of physical experimentation and that this type of technology will increase the technological development of products as experimentation will become faster and more robust before production of any product [19]. As promising as these technological advancements are, there are still concerns with some of the underlying fundamental processes associated with these technologies [20].

Deep learning models have been widely used to generate synthetic data [7]. One such example is a GAN [21]. GANs have been used to generate synthetic image and audio data, especially in creating large training datasets for the use case of facial recognition and speech detection machine learning models [22]. Another example is the inherent bias that is created in visual classification tasks, whereby there exists a balance between the bias and variance generated within the models [23, 24]. As a result, GANs have been shown to work either very well, or not at all at generating useful synthetic data for a specific machine learning task, such as classification using labelled training data [25].

The general architecture of a GAN can be seen below (Fig. 1). As shown in the diagram, a GAN starts with random latent variables about the data it wants to simulate. Thereafter, a generator creates multiple instances of the latent variables based on the observed ranges of the variables. These generated variables are then actuated into a discriminator. At the same time, real data is also moved into the discriminator. The purpose of the discriminator is to differentiate between real and simulated data. Both the simulated and real information is then moved into a condition function. The purpose of the function is to evaluate how well the model can differentiate between the real and the synthetic data. If a data type was classified as fake, then fine tuning on the synthetic data is performed, and the generator updates the synthetic data generation component of the model and moves new synthetic data into the discriminator. This process is repeated for each of the latent variables until the desired amount of data generated through the GAN is achieved [21 – 25].

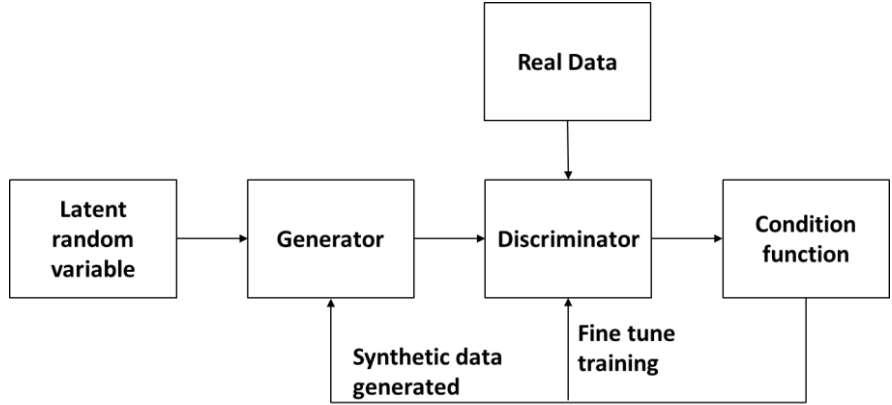

**Fig. 1.** General structure of a Generative Adversarial Network (GAN) adapted from[1]

A BN works with categorical variables where the probabilistic distribution is dependent on the conditional probabilities of a given category within a set of variables. The general probabilistic structure implies that an independent probability is denoted by $P(x)$ and a conditional probability denoted as the $P(x|y_1....,y_i)$. Therefore, $x$ is a function of independent probability, and y a function of conditional probability. As such, probabilities (conditional and independent) can be denoted by the following equation (eq. 1)

$$(P(x_1 \ldots, x_n)) = \prod_{i=1} P(x_i | y(x_i)). \tag{1}$$

Thus, we can apply a BN to a case study where, for example, Exam Level, IQ level, Marks, Aptitude Score leads to a probability of final Admission. If we consider that the following five variables are prevalent within the BN whereby P[$el$] is the probability of Exam Level, P($iql$) the probability of IQ Level, P($as|iql$) is the conditional probability of Aptitude Score given IQ Level, P($m|el,iql$) the conditional probability of Marks given IQ Level and Exam Level, and P($a|m$) the conditional probability of Admission given Marks, then we can create a probabilistic model by which we can calculate the admission score of a candidate (eq. 2)

$$P(el, iql, as, m, a) = P(el) * P(iql) * P(as|iql) * P(m|el, iql) * P(a|m). \tag{2}$$

Once the probabilistic distributions are known, these networks can be represented using a direct acyclic graph [DAG] (Example: Fig. 2).

---

[1] https://www.geeksforgeeks.org/generative-adversarial-network-gan/

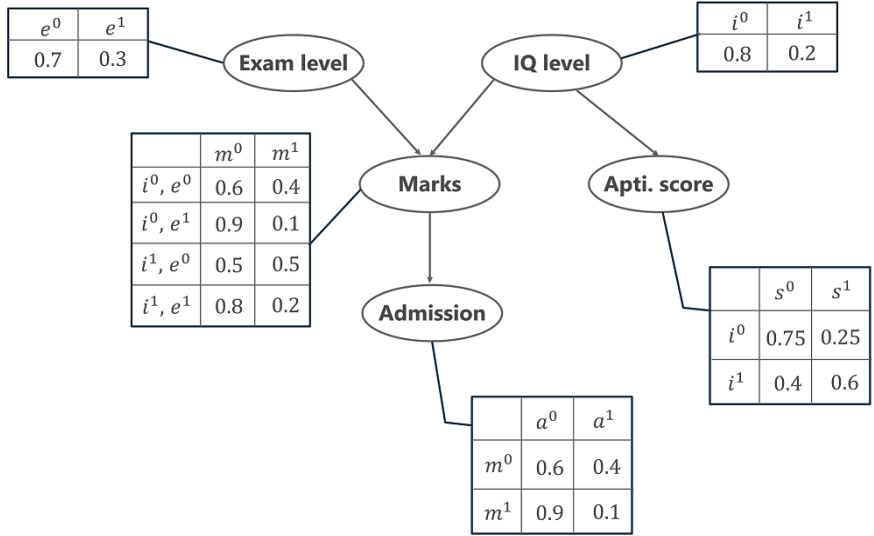

**Fig. 2.** Example of a DAG and probabilistic structure for education[2]

Synthetic data can therefore be generated using different methods and models. Furthermore, simulating tabular data using these approaches is still not widely used in machine learning classification tasks specific to education data. Therefore, the purpose of this article is to compare synthetic tabular data generated from a GAN to those generated from a BN and evaluate the accuracy of this using a machine learning classification task.

## 2  Methodology

Open-source tabular data for the education domain was used in this study[3]. Variables within the dataset were transformed into discrete categorical variables from their original data types[4]. From this data, an expertly defined BN was constructed from the dataset used, and the GAN was applied on the same transformed categorical variables as the BN. The GAN and a BN was used to generate a synthetic dataset for each of the models, containing 10,000 different users' information. The synthetic dataset was generated from an original dataset containing 3029 different samples. The samples as well

---

[2] https://uol.de/en/lcs/probabilistic-programming/webchurch-andopenbugs/example-5-bayesian-network-student-model [last accessed 15 October 2022]

[3] https://github.com/dsfsi/Higher_Education_EDA/tree/main/ opendata [last accessed 15 October 2022]

[4] https://github.com/dsfsi/Higher_Education_EDA/tree/main/synthetic_data_generation/synthetic_data [last accessed 15 October 2022]

as the dataset contained information about the grade point average (GPA) of students spanning over three years of tertiary education, plus one variable summarizing their GPA for high school. For each of the variables in the dataset, the descriptive statistics were compiled [28]. Each of the variables in the synthetic dataset was compared to the original variable within the primary dataset for both the GAN and the BN by looking at their distribution, cumulative sums and density. Furthermore, a machine learning classification algorithm, k-Nearest Neighbour (kNN) was used to predict the target variable in each of the datasets. The kNN algorithm was chosen on the premise of its application to tabular higher education data, but the authors note that several other models could also be used for such an experiment. The kNN algorithm expresses variable features such as Euclidean distances and measures their relevance to one another [29]. A confusion matrix was created based on the classifications made by kNN for each target variable and was further used to evaluate the models for each simulation in the study. A confusion matrix is a means to measure the predicted outcome against the actual outcomes of a given dataset. The results involve comparing the actual ground-truth outcomes from the test dataset to the outcomes predicted based on the trained models. To do so, four primary variables were assessed: true positive results (TP), false positive results (FP), true negative results (TN), and false negative results (FN). Results of the comparison that are labelled as FP are also called type I errors, and outcomes that are listed as FN are also referred to as type II errors. The overall accuracy of the model is the sum of true positive and true negative results divided by the sum of the true results and type I and type II errors (eq. 4)

$$Accuracy = \frac{(TP + TN)}{(TP + FP + TN + FN)} \qquad (4)$$

To illustrate the impact of the underlying data on the performance of the algorithm, a learning curve was used to visualise the results of the prediction classification task. A learning curve is a visualisation technique that shows the performance of a model's accuracy as the training dataset increases. Included in the learning curve is a training score and a cross validation score. A training score can be defined as a measurement of how well the model generalised to fit the training data used. A cross validation score can be defined as a technique used to evaluate kNN model accuracy by training the model on subsets of the training data. The training test split was done on 75% of the synthetic dataset (n = 10000). A total of 10-fold cross validations were used. The authors note that this in not the optimal number of cross fold validations, as the purpose of this experiment was to compare the differences between the two methodologies. The original dataset contained 3029 samples. The synthetic data contained both features and the labels associated with the final outcome of the student university degree.

## 3 Results

The first set of results related to the descriptive statistics of each variable in the dataset showed the differences in the standard deviations (SD) produced by the different models (Table 1).

**Table 1: Descriptive statistics of the variables in the dataset**

| Variable name | Mean BN | SD BN | Mean GAN | SD GAN |
|---|---|---|---|---|
| V1C1 | 0.7055 | 0.455848 | 0.6037 | 0.489153 |
| V1C2 | 0.2945 | 0.455848 | 0.3963 | 0.489153 |
| V2C1 | 0.6844 | 0.464778 | 0.5884 | 0.492148 |
| V2C2 | 0.3156 | 0.464778 | 0.4116 | 0.492148 |
| V3C1 | 0.8001 | 0.399945 | 0.5673 | 0.495475 |
| V3C2 | 0.1999 | 0.399945 | 0.4327 | 0.495475 |
| V4C1 | 0.6917 | 0.461814 | 0.5506 | 0.497458 |
| V4C2 | 0.3083 | 0.461814 | 0.5555 | 0.506372 |

*V = variable

*C = category

As seen from table 1, the overall means of the BN and the GAN differed from one another. Furthermore, the standard deviations of the BN were less than the GAN. This difference implies that the distribution of the data for the BN was more centralized toward the mean of the distribution than the GAN. In figure 3, the distribution of the cumulative sums of each of the four features in the experiment dataset were shown. For the synthetic tabular data generated from the GAN, the overall cumulative sum distribution between the generated and the real data has the highest discrepancy between category 1 and 2 in terms of variable 1, whereas the BN had the highest between 0 and 1. There were slight differences observed between synthetic data produced from GAN and BN.

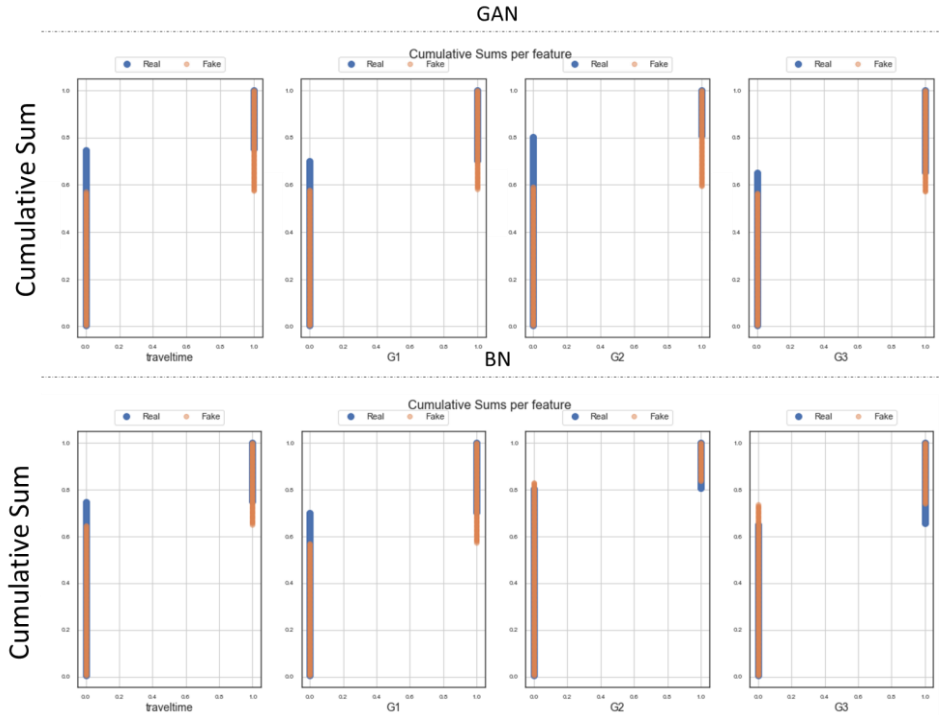

**Fig. 3.** Cumulative sum between GAN and BN for each of the variables.

The density distribution per feature for each of the synthetic datasets were closely related to the primary datasets except where relational interdependence between the variables were important. The parent variables for variable 3 (variable 3 = final mark), were both variable 1 (the first assessment opportunity) and variable 2 (second assessment opportunity). Variable 3 for GAN had the lowest density. Based on these results, the synthetic data generated between the GAN and BN are similar in structure and distribution for each of the variables in terms of the cumulative sums as well as the density distribution (Fig. 4).

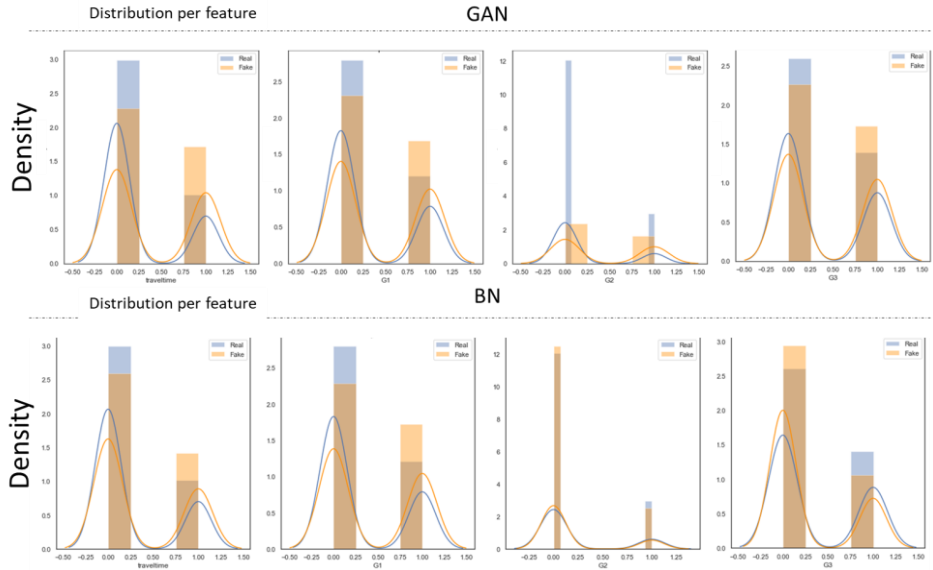

**Fig. 4.** Density distribution of each variable for GAN and BN in the dataset

To illustrate the value of the interdependence that was still maintained between the data generated from the GAN against that of the BN, a classification task was applied on the dataset. The overall accuracy of the classification task for the GAN was at least 20% lower than that of the BN (Fig. 5).

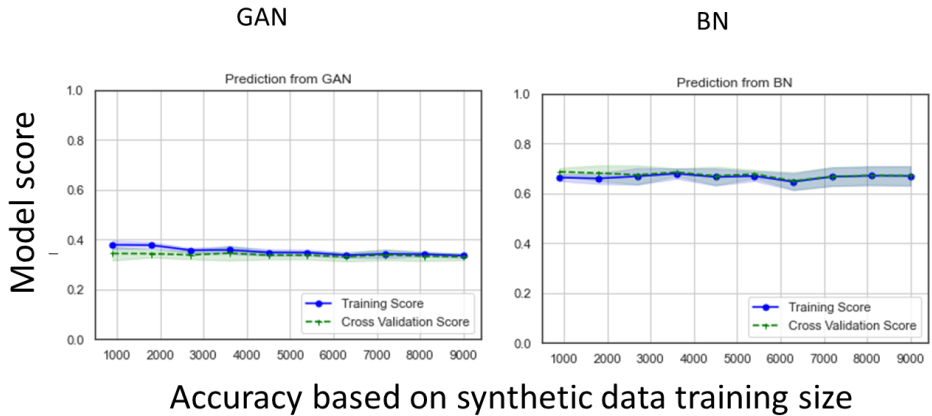

**Fig. 5.** kNN classification task performed on synthetic datasets

Overall, the BN produced not only synthetic data that had lower standard deviations, and a better density distribution between the real and synthetic data, but also produced data that could be used with a relatively high accuracy (62 - 70%) for a classification task. The synthetic data generated from a BN could therefore be used to simulate prediction tasks based on classification algorithms. These results have a variety of potential use cases for education as not only can quantitative data be simulated using this approach, but also processes that require mapping out the student journey, predicting student success rates, and modelling student learning pathways [1, 5, 16].

## 4      Conclusion

Although deep learning architectures are popular for the generation of synthetic data, fundamental probabilistic models have a use case in this regard. Furthermore, if synthetic data needs to be generated for the purpose of machine learning, probabilistic models seem to be a better fit for synthetic tabular data generation than generative deep learning models, such as a GAN. This is because the probabilistic models seem to have higher accuracy scores for classification tasks based on our results. We acknowledge that with complex datasets, a lot of the nuance of what we illustrate in this paper will be lost under the complex interpretation of the latent variables that may or may not be present (as with visual, sound or even text data). Furthermore, we acknowledge that there are a variety of probabilistic and deep learning models that still need to be further explored in terms of their contribution to the creation of synthetic data that display variable associations that are important for this type of variable interdependence. It is also relatively easy to insert expert specific knowledge about a probabilistic distribution. We therefore recommend that other models and other data types be further explored for the creation of synthetic data to identify the limitation of both approaches for synthetic data generation used in education.